\documentclass{article}

\PassOptionsToPackage{numbers}{natbib}
\newif\iffinal
\finaltrue
\iffinal
\usepackage[final]{nips_2016}
\else
\usepackage{nips_2016}
\fi

\makeatletter
\newcommand\footnoteref[1]{\protected@xdef\@thefnmark{\ref{#1}}\@footnotemark}
\makeatother

\usepackage[utf8]{inputenc} 
\usepackage[T1]{fontenc}    
\usepackage{hyperref}       
\usepackage{url}            
\usepackage{booktabs}       
\usepackage{amsfonts}       
\usepackage{amsmath}
\usepackage{nicefrac}       
\usepackage{microtype}      
\usepackage{graphicx}
\usepackage{sparklines}

\DeclareMathOperator\Cat{Cat}

\title{AutoMOS: Learning a non-intrusive assessor of naturalness-of-speech}

\author{
  Brian~Patton \\
  \texttt{bjp@google.com} \\
  \And
  Yannis~Agiomyrgiannakis \\
  \texttt{agios@google.com} \\
  \And
  Michael~Terry \\
  \texttt{michaelterry@google.com} \\
  \And
  Kevin~Wilson \\
  \texttt{kwwilson@google.com} \\
  \And
  Rif A.~Saurous \\
  \texttt{rif@google.com} \\
  \And
  D.~Sculley \\
  \texttt{dsculley@google.com} \\
}

\begin{document}

\maketitle

\begin{abstract}
  Developers of text-to-speech synthesizers (TTS) often make use of
  human raters to assess the quality of synthesized speech. We
  demonstrate that we can model human raters' mean opinion scores
  (MOS) of synthesized speech using a deep recurrent neural network
  whose inputs consist solely of a raw waveform. Our best models
  provide utterance-level estimates of MOS only moderately inferior to
  sampled human ratings, as shown by Pearson and Spearman
  correlations. When multiple utterances are scored and averaged,
  a scenario common in synthesizer quality assessment,
  AutoMOS achieves correlations approaching those of human raters.
  The AutoMOS model has a number of applications, such as the
  ability to explore the parameter space of a speech
  synthesizer without requiring a human-in-the-loop.
\end{abstract}

\section{Introduction}

To evaluate changes to text-to-speech (TTS) synthesizers, human
raters are often employed to assess the synthesized
speech. Multiple human ratings of an audio sample contribute 
to a \textit{mean opinion score} (MOS). MOS has been crowdsourced 
with specific attention to rater quality by \citep{ribeiro2011crowdmos}. 
While crowdsourcing introduces a degree of parallelism to the rating 
process, it is still relatively costly and time-consuming to obtain MOS for 
TTS quality testing.

Numerous systems have been produced to algorithmically
produce \textit{objective} assessments of audio quality approximating
the \textit{subjective} human assessment, including some which assess 
speech quality. 
For example, MCD \citep{kubichek1993mel}, 
PESQ \citep{rix2001perceptual} and
POLQA \citep{rec2011p} target this particular space. 
These are \textit{intrusive} assessors in that they assume the 
presence of an undistorted reference signal to facilitate
comparisons when deriving ratings, something that does not exist in
the case of the synthesized speech of TTS.
\textit{Non-intrusive} assessments such as ANIQUE \citep{kim2005anique},
LCQA \citep{grancharov2006non} and P.~563 \citep{malfait2006p} have been proposed 
to evaluate speech quality where this reference signal is not available.
Much research in quality assessment is targeted at telephony, with
emphasis on detecting distortions and other artifacts introduced by 
lossy compression and transmission.

Throughout this work, a \textit{synthesizer} constitutes 
a snapshot of the evolving implementation of a unit selection synthesis 
algorithm and a continually growing corpus of recorded audio,
combined with a specific set of synthesis/cost parameters. When we 
partition or aggregate data by synthesizer, 
we take all utterances from a given synthesizer 
and allocate them \textit{en masse} to a single training or evaluation fold, 
or to a single aggregate metric, e.g. \textit{synthesizer-level mean MOS}.

In \citep{peng2002perpetually}, 
the authors used human ratings to improve the correlation between MOS 
and unit selection TTS cost. Similarly, \citep{alias2011efficient, pobar2012optimization} 
explore means of tuning cost functions to incorporate subjective preferences.
Such works consider direct optimization of synthesizer MOS as a function of 
synthesizer parameters (e.g. cost function weights). In prior unpublished work we 
trained similar models and found they could exceed 0.9 Spearman rank correlation 
between true and estimated synthesizer MOS.
However, any modifications to parameter semantics or engine internals render this 
mapping invalid. It is desirable to learn a synthesizer assessment which operates
independently of engine internals, directly assessing pools of TTS waveforms.

We demonstrate that deep recurrent networks can model 
\textit{naturalness}-of-speech MOS ratings
produced by human raters for TTS synthesizer evaluation, 
using only raw audio waveform as input. We explore a variety of deep recurrent
architectures, to incorporate long-term time dependencies.
Our tuned AutoMOS model achieves a Spearman rank correlation of 0.949, ranking 36 different 
synthesizers. (Sampling a single human rating for 
each utterance yields a Spearman correlation of 0.986.)
When evaluating the calibration of AutoMOS on multiple utterances with 
similar predicted MOS, we find five-fold median correlations > 0.9 and 
MSE competitive with sampled human ratings, even when we quantize the predicted 
utterance MOS to the 0.5 increments of the human rater scale.
Such results open the door for scalable, automated tuning and continuous 
quality monitoring of TTS engines.

\section{Model \& Results}

Because audio data is of varying length for each example, either 
directly pooling across the time dimension or the use of 
recurrent neural networks (RNN) is suggested.
To encode the intuition that valuable information exists at relatively large time-scales 
(consider phone or transition duration or inflection, which may vary according to context),
we opt to explore a family of RNN models.

In particular, we test a family of models that layer one or more fully-connected 
layers atop the time-pooled outputs of a stack of recurrent Long Short-Term Memory 
\citep[LSTM;][]{hochreiter1997long} cells. The timeseries input to the LSTM is
either a log-mel spectrogram or a time-pooled convolution as in \citep{hoshen2015speech}, 
in each case over a 16kHz waveform.
We consider the addition of single frame velocity and acceleration components 
to this timeseries.
The final LSTM layer's outputs are max-pooled
across time, and fed as inputs to the fully-connected
hidden layers which compute final regression values.
We explored a means of inducing learning across longer timeframes 
(a stacked LSTM where deeper layers use a stride of 2 or more timesteps 
over the outputs of of lower-level LSTM layers), but found performance 
comparable to that of a simpler stacked or single-layer LSTM. 
Max-pooling non-final LSTM layers' outputs and adding skip connections 
to the final hidden layers was not found to improve performance.

\begin{figure}
  \includegraphics[width=\linewidth]{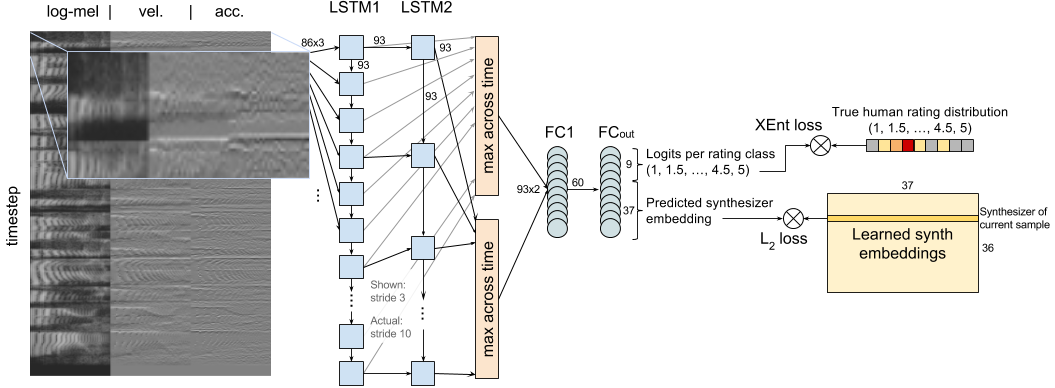}
  \caption{Diagram of the best performing AutoMOS network}
  \label{fig:diagram}
\end{figure}

We explore multiple modes of predicting and training over input waveforms $x$: 
(1) predict sufficient statistics $\mu(x), \sigma(x)$ and 
train on log-likelihood of individual human ratings $r_i$ under
Gaussian $\log p(r_i | \mu(x), \sigma(x))$, 
(2) predict $MOS(x)$ and train on utterance-level L2 loss $(MOS(x) - MOS_{true})^2$, and 
(3) predict $logits(x)$ and train on cross-entropy between the true 9-category distribution 
of human ratings and $\Cat(logits(x))$ per-utterance.
We train a separate set of outputs with a learned embedding of the ground-truth 
synthesizer, providing both regularization and gradients to the training process. 
Embeddings are initialized randomly; both the embedding and the prediction thereof 
receive a gradient (toward each other) in each training step. 
The best performing model is illustrated in Figure~\ref{fig:diagram}.

All models were trained with Adagrad on batches of 20 examples asynchronously across 10 workers.
We use five-fold cross-validation to evaluate the best found set of hyperparameters, with 
all utterances for any given synthesizer appearing exclusively in a single fold.

\paragraph{Data}
We use a corpus of TTS naturalness scores acquired over multiple years 
across multiple instances of quality testing for 
\iffinal Google's \else [[our]] \fi 
TTS engines. All tests are iterations on a single
English (US) voice used across multiple products. Raters scored each utterance given 
a 5-point Likert scale for naturalness, in half-point increments.
We partition training data from holdout data such that all utterances for a given synthesizer
are in the same partition.
The data includes 168,086 ratings across 47,320 utterances generated by 36 synthesizers. The 
utterance quantity per synthesizer varies from 64 to 4800: \begin{sparkline}{14}
\setlength{\sparkspikewidth}{1pt}
\definecolor{sparkspikecolor}{named}{blue}
\definecolor{sparkbottomlinecolor}{gray}{0.9}
\sparkspike 0.027027 0.272349  
\sparkspike 0.054054 0.347564  
\sparkspike 0.081081 0.383818  
\sparkspike 0.108108 0.383818  
\sparkspike 0.135135 0.383818  
\sparkspike 0.162162 0.383818  
\sparkspike 0.189189 0.464384  
\sparkspike 0.216216 0.464384  
\sparkspike 0.243243 0.464384  
\sparkspike 0.270270 0.464384  
\sparkspike 0.297297 0.464384  
\sparkspike 0.324324 0.464384  
\sparkspike 0.351351 0.464384  
\sparkspike 0.378378 0.464384  
\sparkspike 0.405405 0.464384  
\sparkspike 0.432432 0.532720  
\sparkspike 0.459459 0.581204  
\sparkspike 0.486486 0.581204  
\sparkspike 0.513514 0.581204  
\sparkspike 0.540541 0.581204  
\sparkspike 0.567568 0.581204  
\sparkspike 0.594595 0.581204  
\sparkspike 0.621622 0.618812  
\sparkspike 0.648649 0.618812  
\sparkspike 0.675676 0.618812  
\sparkspike 0.702703 0.675520  
\sparkspike 0.729730 0.689380  
\sparkspike 0.756757 0.698025  
\sparkspike 0.783784 1.000000  
\sparkspike 0.810811 1.000000  
\sparkspike 0.837838 1.000000  
\sparkspike 0.864865 1.000000  
\sparkspike 0.891892 1.000000  
\sparkspike 0.918919 1.000000  
\sparkspike 0.945946 1.000000  
\sparkspike 0.972973 1.000000  
\sparkbottomline 0.5
\end{sparkline}, in log-scale.

\paragraph{Hyperparameter Tuning}
We used Google Cloud's HyperTune to explore a set of hyperparameters shown in 
Table~\ref{hparams-table}.
About 2 in 3 top-performing tuning runs used the cross-entropy categorical training mode.
The top 10 configurations we found had eval-set 
Pearson correlations between utterance-level predicted and true MOS 
ranging from 0.56-0.61 after 20,000 training steps.
When we constrained the search space to 
those models using convolution+pooling based timeseries (as opposed to log-mel), we found
weaker best eval-set correlations around 0.48. This could indicate there is little 
value in the sample-level details when dealing in synthesized speech, or could 
signal insufficient training data.
While \citep{sainath2015learning} reported gammatone-like learned filter banks, 
their speaker-independent ASR covers a much wider range of voices, 
and we did not observe a similar set of emergent filters from random initialization.
Initialization with gammatone filters yielded only nominal improvements in performance (r=0.51).

Relative to a simple L2 loss to the true MOS, we observed little benefit from
training a Gaussian predictor against individual ratings. 
The estimated variance was typically higher than the true sample variance 
for a given utterance. Using a simpler L2 loss against the true MOS provided 
for faster training and convergence, allowing us to try a wider 
variety of structural changes. Treating predictions as categorical and using a cross-entropy
loss slightly outperformed the L2 construction in the top tuning runs (0.61 categorical vs. 0.58 L2).
The categorical form gives AutoMOS more weights and hence a greater capacity
near the output layer.

\begin{table}[t]
  \caption{Model Hyperparameters}
  \label{hparams-table}
  \centering
  \begin{tabular}{lll}
    \toprule
    Description                      & Range explored       & Best Performer \\
                                     &                      & 
                                                \begin{small}(Pearson r = 0.61)\end{small} \\
    \midrule
    Learning rate; decay / 1000 steps        & 0.0001 - 0.1; 0.9 - 1.0    & 0.057; 0.94\\
    L1; L2 regularization                    & 0.0 - 0.001            & 1.4e-5; 2.6e-5\\
    Loss strategy                          & L2 $|$ cross-entropy   & cross-entropy \\
    Synthesizer regression embedding dim      & 0 - 50             & 37 \\
    Timeseries type                          & log-mel $|$ pooled conv1d & log-mel \\
    Timeseries width (\# mel bins, conv filters) & 20 - 100  & 86\\
    Timeseries 1-step derivatives          & (none) $|$ vel. $|$ vel. + acc.   & vel. + acc. \\
    LSTM layer width; depth                 & 20 - 100; 1 - 10      & 93; 2 \\
    LSTM timestep stride at non-0th layers & 1 - 10           & 10 \\
    LSTM layers feeding hidden layer inputs & all $|$ last    & all \\
    Post-LSTM hidden layer width; depth     & 20 - 200; 0 - 2 & 60; 1 \\
    \bottomrule
  \end{tabular}
\end{table}

\paragraph{Evaluation}
As simple baselines for comparison, we consider (1) a bias-only model which always predicts the mean of all
observed utterances' MOS and (2) a small nonlinear model which takes only utterance length as input
(two 10-unit hidden layers with rectified linear activation),
with the intuition that a longer utterance includes more opportunities to make mistakes 
deemed unnatural. We draw one human rating for each utterance and show this comparison as a
"Sample human rating" column.

\begin{figure}
  \includegraphics[width=.74\linewidth]{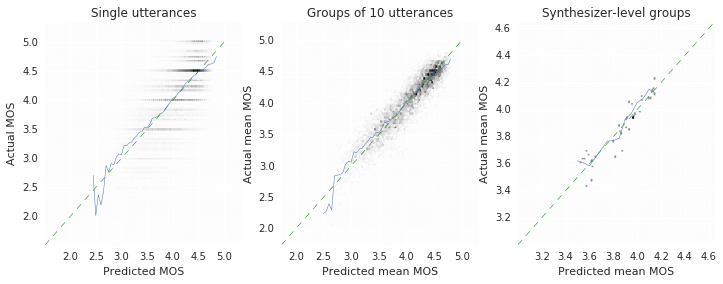}
  \includegraphics[width=.25\linewidth]{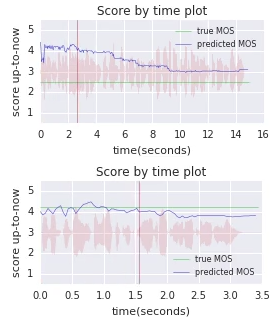}
  \caption{\textbf{Left:} Calibration plots (including all eval folds of AutoMOS): Green represents perfect calibration; blue plots $(\mathbb{E}_{w}(MOS_{predicted})$, $\mathbb{E}_{w}(MOS_{true}))$ within 0.05 windows $w$ along the x-axis. \textbf{Right:}~Samples from score-over-time animations. Visit \texttt{\href{http://goo.gl/cnQbSn}{goo.gl/cnQbSn}} to view.}
  \label{fig:calibplots}
\end{figure}

If errors are unbiased, an increased sample size should reduce error.
We sort utterances by predicted MOS 
and evaluate correlations between 
$ \mathbb{E}_{group}(MOS_{predicted}) $ and $ \mathbb{E}_{group}(MOS_{true}) $ 
(where $\mathbb{E}$ is the expected value operator)
on groupings of 10 
 or more utterances with adjacent predicted MOS in a 
similar fashion to a calibration plot. 
 We show such plots in Figure~\ref{fig:calibplots}.

How well can we rank synthesizers relative to one another? 
To perform this evaluation, we use the above five-fold cross validation 
to predict a MOS for each utterance using the AutoMOS instance from which it was held-out. 
We then average at the synthesizer-level, giving us a total of 36 
$ \mathbb{E}_{synth}(MOS_{predicted}), \mathbb{E}_{synth}(MOS_{true})  $ 
pairs upon which we evaluate. Results shown in Table~\ref{results-table}. 

\begin{table}[t]
  \caption{RMSE and correlation results (reflecting median fold, except as indicated)}
  \label{results-table}
  \centering
  \begin{minipage}{\linewidth}
  \renewcommand\footnoterule{ \kern -1ex}
  \begin{tabular}{lccccc}
    \toprule
    & \multicolumn{2}{c}{Baselines} & \multicolumn{2}{c}{AutoMOS} & Ground-truth \\
    \cmidrule(lr{.5em}){2-3}     \cmidrule(lr{.5em}){4-5}    \cmidrule(lr{.5em}){6-6}
    Metric / Model & Bias-only & NNet(utt. length) & Raw & Quantized\footnote{\label{note1}utterance scores from 1-5 in increments of 0.5} & Sample human rating\footnoteref{note1} \\
    \midrule
    \multicolumn{5}{l}{Utterance-level ($n_{fold} = {6000, 6424, 6624, 12348, 15924}$)} \\ 
    ~~RMSE             & 0.618      & 0.553 & 0.462 & 0.483 & 0.512 \\ 
    ~~Pearson r        & $-$        & 0.454 & 0.668 & 0.638 & 0.764 \\ 
    ~~Spearman r       & $-$        & 0.399 & 0.667 & 0.636 & 0.757 \\   
    \multicolumn{5}{l}{10 utterance means ($n_{fold} = {600, 643, 663, 1235, 1593}$)} \\ 
    ~~RMSE             & 0.203      & 0.213 & 0.172 & 0.171 & 0.358 \\ 
    ~~Pearson r        & $-$        & 0.812 & 0.930 & 0.933 & 0.962 \\ 
    ~~Spearman r       & $-$        & 0.657 & 0.925 & 0.925 & 0.956 \\   
    \multicolumn{5}{l}{Synthesizer-level means ($n = {36}$; uses all folds)} \\ 
    ~~RMSE             & 0.252      & 0.132 & 0.073 & 0.075 & 0.034 \\ 
    ~~Pearson r        & $-$        & 0.795 & 0.938 & 0.935 & 0.987 \\ 
    ~~Spearman r       & $-$        & 0.679 & 0.949 & 0.947 & 0.986 \\   
    \bottomrule
  \end{tabular}
  \end{minipage}
\end{table}

\section{Discussion}

AutoMOS tends to avoid very-high or very-low predictions, likely
reflecting the distribution of the training data. It also seems to 
learn patterns in the data 
around certain common "types" of utterances which usually
achieve high ("OK, setting your alarm") or low MOS (reading dictionary definitions).
It's possible that different distributions of texts per synthesizer 
could yield easily predictable differences in synthesizer-MOS.
A future improvement would be predicting MOS for the raw 
text and evaluating the \textit{advantage} of an utterance relative to 
this baseline. In \citep{peng2002perpetually}, naturalness is predictable from
unit selection costs; here, we want to remove the predictive 
baseline of the text.

We have begun tuning a TTS engine using AutoMOS. 
Subsequent human evaluations will provide concrete results on the 
model and evaluation criteria we've selected. 
Similarly, we will experiment with the system for 
continuous quality testing of a large-scale TTS deployment.
It may be possible to leverage AutoMOS to do stratified sampling of utterances to send to
human raters. This would allow raters to focus energy more evenly across the quality 
spectrum.

To probe what's been learned, we have explored artificial truncation, as in Figure~\ref{fig:calibplots} (right). Methods like layerwise relevance propagation \citep{binder2016layer} 
or activation difference propagation \citep{shrikumar2016not} have
shown promise with image models, and could be interesting to apply to a unit selection cost function.

\newpage
\begin{small}
\bibliographystyle{apalike}
\bibliography{paper} 

\begin{thebibliography}{}

\bibitem[Al{\'\i}as et~al., 2011]{alias2011efficient}
Al{\'\i}as, F., Formiga, L., and Llora, X. (2011).
\newblock Efficient and reliable perceptual weight tuning for unit-selection
  text-to-speech synthesis based on active interactive genetic algorithms: A
  proof-of-concept.
\newblock {\em Speech Communication}, 53(5):786--800.

\bibitem[Binder et~al., 2016]{binder2016layer}
Binder, A., Bach, S., Montavon, G., M{\"u}ller, K.-R., and Samek, W. (2016).
\newblock Layer-wise relevance propagation for deep neural network
  architectures.
\newblock In {\em Information Science and Applications (ICISA) 2016}, pages
  913--922. Springer.

\bibitem[Grancharov et~al., 2006]{grancharov2006non}
Grancharov, V., Zhao, D.~Y., Lindblom, J., and Kleijn, W.~B. (2006).
\newblock Non-intrusive speech quality assessment with low computational
  complexity.
\newblock In {\em INTERSPEECH}.

\bibitem[Hochreiter and Schmidhuber, 1997]{hochreiter1997long}
Hochreiter, S. and Schmidhuber, J. (1997).
\newblock Long short-term memory.
\newblock {\em Neural computation}, 9(8):1735--1780.

\bibitem[Hoshen et~al., 2015]{hoshen2015speech}
Hoshen, Y., Weiss, R.~J., and Wilson, K.~W. (2015).
\newblock Speech acoustic modeling from raw multichannel waveforms.
\newblock In {\em 2015 IEEE International Conference on Acoustics, Speech and
  Signal Processing (ICASSP)}, pages 4624--4628. IEEE.

\bibitem[ITU-T, 2011]{rec2011p}
ITU-T (2011).
\newblock P. 863, {P}erceptual objective listening quality assessment
  ({POLQA}).
\newblock {\em International Telecommunication Union, CH-Geneva}.

\bibitem[Kim, 2005]{kim2005anique}
Kim, D.-S. (2005).
\newblock {ANIQUE}: An auditory model for single-ended speech quality
  estimation.
\newblock {\em IEEE Transactions on Speech and Audio Processing},
  13(5):821--831.

\bibitem[Kubichek, 1993]{kubichek1993mel}
Kubichek, R. (1993).
\newblock Mel-cepstral distance measure for objective speech quality
  assessment.
\newblock In {\em Communications, Computers and Signal Processing, 1993., IEEE
  Pacific Rim Conference on}, volume~1, pages 125--128. IEEE.

\bibitem[Malfait et~al., 2006]{malfait2006p}
Malfait, L., Berger, J., and Kastner, M. (2006).
\newblock P. 563: {T}he {ITU-T} standard for single-ended speech quality
  assessment.
\newblock {\em IEEE Transactions on Audio, Speech, and Language Processing},
  14(6):1924--1934.

\bibitem[Peng et~al., 2002]{peng2002perpetually}
Peng, H., Zhao, Y., and Chu, M. (2002).
\newblock Perpetually optimizing the cost function for unit selection in a
  {TTS} system with one single run of {MOS} evaluation.
\newblock In {\em INTERSPEECH}.

\bibitem[Pobar et~al., 2012]{pobar2012optimization}
Pobar, M., Martincic-Ipsic, S., and Ipsic, I. (2012).
\newblock Optimization of cost function weights for unit selection speech
  synthesis using speech recognition.
\newblock {\em Neural Network World}, 22(5):429.

\bibitem[Ribeiro et~al., 2011]{ribeiro2011crowdmos}
Ribeiro, F., Flor{\^e}ncio, D., Zhang, C., and Seltzer, M. (2011).
\newblock Crowdmos: An approach for crowdsourcing mean opinion score studies.
\newblock In {\em 2011 IEEE International Conference on Acoustics, Speech and
  Signal Processing (ICASSP)}, pages 2416--2419. IEEE.

\bibitem[Rix et~al., 2001]{rix2001perceptual}
Rix, A.~W., Beerends, J.~G., Hollier, M.~P., and Hekstra, A.~P. (2001).
\newblock Perceptual evaluation of speech quality ({PESQ})-a new method for
  speech quality assessment of telephone networks and codecs.
\newblock In {\em Acoustics, Speech, and Signal Processing, 2001.
  Proceedings.(ICASSP'01). 2001 IEEE International Conference on}, volume~2,
  pages 749--752. IEEE.

\bibitem[Sainath et~al., 2015]{sainath2015learning}
Sainath, T.~N., Weiss, R.~J., Senior, A., Wilson, K.~W., and Vinyals, O.
  (2015).
\newblock Learning the speech front-end with raw waveform {CLDNNs}.
\newblock In {\em Proc. Interspeech}.

\bibitem[Shrikumar et~al., 2016]{shrikumar2016not}
Shrikumar, A., Greenside, P., Shcherbina, A., and Kundaje, A. (2016).
\newblock Not just a black box: Learning important features through propagating
  activation differences.
\newblock {\em arXiv preprint arXiv:1605.01713}.

\end{thebibliography}
\end{small}

\end{document}